\documentclass[10pt,twocolumn,letterpaper]{article}

\usepackage{cvpr}
\usepackage{times}
\usepackage{epsfig}
\usepackage{graphicx}
\usepackage{amsmath}
\usepackage{amssymb}
\usepackage{authblk}

\newcommand{\sF}{\mathcal{F}}
\newcommand{\bO}{\mathbf{O}}
\newcommand{\bI}{\mathbf{I}}
\newcommand{\bD}{\mathbf{D}}
\newcommand{\sS}{\mathcal{S}}
\newcommand{\sP}{\mathcal{P}}
\newcommand{\sQ}{\mathcal{Q}}
\newcommand{\sV}{\mathcal{V}}
\newcommand{\transpose}{^\mathsf{T}}
\newcommand{\btheta}{\boldsymbol{\theta}}

\usepackage[breaklinks=true,bookmarks=false,colorlinks]{hyperref}

\cvprfinalcopy 

\def\cvprPaperID{3004} 

\ifcvprfinal\fi
\begin{document}

\title{Channel Attention based Iterative Residual Learning for Depth Map Super-Resolution}

\author{Xibin Song$^{1,2}$, Yuchao Dai$^3$\thanks{Corresponding author}, Dingfu Zhou$^{1,2*}$, Liu Liu$^{5,6}$, Wei Li$^{4}$, Hongdong Li$^{5,6}$ \\ and Ruigang Yang$^{1,2,7}$}

\affil{$^1$Baidu Research \quad $^2$National Engineering Laboratory of Deep Learning Technology and Application, China \quad $^3$ Northwestern Polytechnical University, China \quad $^4$ Shandong University, China \quad \\
$^5$ Australian National University, Australia \quad $^6$Australian Centre for Robotic Vision, Australia \quad
$^7$University of Kentucky, Kentucky, USA \\
\tt\small \{songxibin,zhoudingfu\}@baidu.com, daiyuchao@gmail.com}

\maketitle
\thispagestyle{empty}


\begin{abstract}
Despite the remarkable progresses made in deep-learning based depth map super-resolution (DSR), how to tackle real-world degradation in low-resolution (LR) depth maps remains a major challenge.
Existing DSR model is generally trained and tested on synthetic dataset, which is very different from what would get from a real depth sensor. 
In this paper, we argue that DSR models trained under this setting are restrictive and not effective in dealing with real-world DSR tasks. We make two contributions in tackling real-world degradation of different depth sensors. 
First, we propose to classify the generation of LR depth maps into two types: non-linear downsampling with noise and interval downsampling, for which DSR models are learned correspondingly. Second, we propose a new framework for real-world DSR, which consists of four modules : 1) An iterative residual learning module with deep supervision to learn effective high-frequency components of depth maps in a coarse-to-fine manner; 2) A channel attention strategy to enhance channels with abundant high-frequency components; 3) A multi-stage fusion module to effectively re-exploit the results in the coarse-to-fine process; and 4) A depth refinement module to improve the depth map by TGV regularization and input loss.
Extensive experiments on benchmarking datasets demonstrate the superiority of our method over current state-of-the-art DSR methods.
\end{abstract}


\section{Introduction}
Depth maps have been widely embraced as a new technology by providing complementary information in many applications~\cite{hartley2012efficient}\cite{liu:efficient:cvpr2017}\cite{song:edge-guided-depth-map-enhancement:icpr2016}\cite{song2019apollocar3d}\cite{song:mtap:2018}\cite{song:estimation-of-kinect-depth-confidence-through-self-training:cgi2014}. However, depth sensors, such as Microsoft Kinect and Lidar, can only provide depth maps of limited resolutions. Hence, depth map super-resolution (DSR) draws more and more attentions. As a fundamental low-level vision problem, DSR aims at super-resolving a high-resolution (HR) depth map from a low-resolution (LR) depth map input ~\cite{riegler:A-deep-primal-dual-netwok-for-guided-depth-super-resolution:BMVC2016}\cite{hui:Depth-map-super-resolution-by-deep-multi-scale-guidance:ECCV2016}\cite{Riegler:ATGV-Net:ECCV2016}\cite{song-Deep-depth-super-resolution:ACCV2016}\cite{song-Deeply-supervised-depth-map-super-resolution:TCSVT2019}\cite{Voynov:perceptual-deep-depthsr:2019_ICCV}, which is a challenging task due to the great information loss in the down-sampling process. Besides, depth maps generally contain less textures and more sharp boundaries, and are usually degraded by noise due to the imprecise consumer depth cameras, which further increase the challenge.

\begin{figure}
  \centering
  \includegraphics[width=\linewidth]{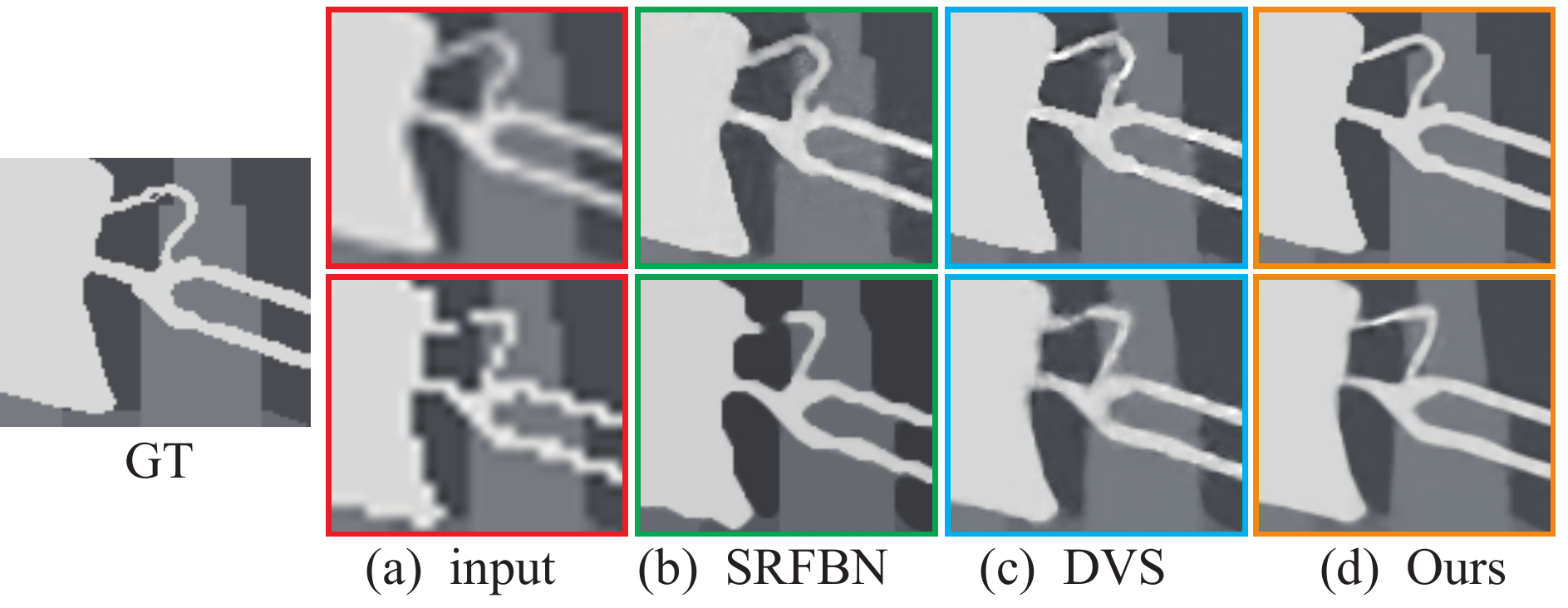}
  \caption{Results of different methods using different types of LR depth maps as input ($\times 4$). (a) input, (b) SRFBN~\cite{Li-SRFBN:CVPR2019}, (c) DVS~\cite{song-Deeply-supervised-depth-map-super-resolution:TCSVT2019} and (d) Ours. The first row shows the results under non-linear (bi-cubic) down-sampling degradation, while the second row shows the results under interval down-sampling degradation.}
  \label{fig:first_figure}
\end{figure}

Recently, significant progress has been made in super-resolution by using convolutional neural networks (CNNs) in regression ways, both in color image super-resolution (CSR) and DSR \cite{Learning-a-deep-convolutional-network-for-image-super-resolution:ECCV-2014}\cite{riegler:A-deep-primal-dual-netwok-for-guided-depth-super-resolution:BMVC2016}\cite{Hu_Meta-SR:2019_CVPR}\cite{hui:Depth-map-super-resolution-by-deep-multi-scale-guidance:ECCV2016}\cite{Li-SRFBN:CVPR2019}\cite{Liu:ABPN-ICCVW2019}\cite{Voynov:perceptual-deep-depthsr:2019_ICCV}. 
These methods usually apply bi-cubic downsampling as the degradation model and add noise to simulate the generation of LR images. Besides,~\cite{Gu_blendsr:2019_CVPR} and~\cite{Zhang_deep_plug_and_play_sr:2019_CVPR} propose to estimate the down-sampling kernels to estimate the degradation of LR images. However, bi-cubic degradation model and degradation kernels are insufficient to describe the process of depth map down-sampling. 

Depth map exists in different types in real world, which can be classified into two types: (1). depth maps with smoothed surfaces, such as depth maps generated by stereo matching~\cite{Barnard:Stereo}\cite{Marr:Stereo:Science}\cite{Okutomi:multi-basedline-stereo:PAMI} and depth maps captured by low-cost sensors (Kinect); (2). depth maps with sharp boundaries, such as depth maps captured by Lidar. For (1), depth maps are always smooth, thus, non-linear downsampling degradation model and down-sampling kernels can be used to simulate the generation of LR depth maps. For (2), depth maps captured by Lidar are generated from 3D points of real world. They are always with sharp boundaries. Imaging the projection process of 3D points onto a 2D image, when two 3D points are projected to a same 2D coordinates in a depth map, it should reserve the 3D point with smaller depth $z$ due to occlusion. Interpolation (bi-cubic or degradation kernel) is not suitable in such process, hence we argue that bi-cubic degradation and blur kernels are not reasonable, and we propose to use interval down-sampling degradation to describe the down-sampling progress. Fig.~\ref{fig:first_figure} (a) illustrates the two types of LR depth maps, where interval down-sampling and non-linear degradation have quite different manifestations.

In this paper, to effectively tackle the two types of depth maps (non-linear degradation with noise and interval down-sampling degradation), we adopt an iterative residual learning framework with deep supervision (coarse-to-fine), which guarantees that each sub-module can gradually obtain the high-frequency components of depth maps step by step. Besides, in each sub-module, channel attention strategy is utilized to enhance the channels with more effective information, thus, obtains better results. What's more, the inter-media results obtained by different sub-modules are fused to provide effective information to tackle different types of depth maps. Total Generalized Variation (TGV) term and input loss are utilized to further refine the obtained HR depth maps. Any support of HR color information is not needed and weight sharing between different sub-modules can effective reduce the number of parameters, which makes our proposed approach much more flexible. The proposed framework is trained in an end-to-end manner, and experiments on various benchmarking datasets demonstrate the superiority of our method over state-of-the-art super-resolution methods, including both DSR and CSR methods.


Our main contributions are summarized as:
\begin{itemize}
\vspace{-2mm}
\item To tackle real world degradation in low-resolution depth maps, we propose to classify the generation of LR depth maps into two types: non-linear down-sampling with noise and interval downsampling, for which DSR models are learned correspondingly.

\vspace{-2mm}
\item We propose an iterative residual learning based framework for real world DSR, where channel attention, multi-stage fusion, weight sharing and depth refinement are employed to learn HR depth maps in a coarse-to-fine manner.

\vspace{-2mm}
\item 
Extensive experiments on various benchmarking datasets  demonstrate  the  superiority  of  our proposed framework over current state-of-the-art DSR methods.


\end{itemize}

\section{Related work}
In this section, we briefly review related work in both color image super-resolution (CSR) and depth map super-resolution (DSR). 

\subsection{DCNN based CSR}

In CSR, bi-cubic down-sampling degradation are commonly used down-sampling methods to generate LR color images. Methods, such as
~\cite{Learning-a-deep-convolutional-network-for-image-super-resolution:ECCV-2014}\cite{Kim:Accurate-image-super-resolution-using-very-deep-convolutional-networks:CVPR2016}\cite{Tai:image-super-resolution-via-deep-recursive-residual-network:cvpr2017}\cite{Sajjadi:enhancement:single-image-SR-through-automated-texture-synthesis:iccv2017}, have proven that CNN outperformed conventional learning approaches with large margin. These methods regard super-resolution as an LR color image to HR color image regression problem, and generate an end-to-end mapping between LR and HR image. Besides, residual architectures, such as~\cite{Cai_2019_ICCV}\cite{Hu_Meta-SR:2019_CVPR}\cite{Qiu_2019_ICCV:residual-network}\cite{zhang:Residual-dense-network-for-imageSR:cvpr2018} are commonly used in solving CSR. HR color images are generated by learning the residuals between LR images and groundtruth. Recently, back projection strategy, such as~\cite{Deng_2019_ICCV}\cite{He_OED_net_sr:2019_CVPR}\cite{Li-SRFBN:CVPR2019}\cite{Liu:ABPN-ICCVW2019}, are proved to have well performance by representing the LR and HR feature residuals in more efficient ways. Meanwhile, attention based model is also utilized in CSR. To obtain more discriminative representations,~\cite{Dai_second-order-sr-2019_CVPR}\cite{channel_attention_networks:eccv2018} propose to use attention strategy to enhance feature representation. Kernel based methods~\cite{Gu_blendsr:2019_CVPR} \cite{Zhang_deep_plug_and_play_sr:2019_CVPR} \cite{Zhou_2019_ICCV:kernel} are also utilized in CSR, which estimate a blur kernel to simulate the generation of LR color images. Besides, \cite{Lutio_2019_ICCV:pixel-to-pixel-transformation} exploits pixel to pixel transfer techonolgy in solving the problem of CSR.

\subsection{Depth Map Super-resolution}

\subsubsection{Conventional Learning based DSR} 
To solve the problem of DSR, prior information is used as useful guidance to generate HR depth maps from LR depth maps. Using prior information learned from additional depth map datasets, ~\cite{Depth-super-resolution-by-rigid-body-self-similarity-in-3d:CVPR-2013}\cite{Patch-based-synthesis-for-single-depth-image-super-resolution:ECCV-2012}\cite{Xie:edge-guided-single-depth-sr:TIP2015} propose to use MRF method to solve the problem of DSR. Meanwhile, other learning based methods, such as sparse representation and dictionary learning, are utilized in DSR.~\cite{Varitional-depth-superresolution-using-example-based-edge-representations:ICCV-2015} proposes to exploit sparse coding strategy and Total Generalized Variation (TGV) to effectively generate HR depth edges, which are used as useful guidance in the generation of HR depth maps. Besides, using HR color image as effective guidance, ~\cite{Image-guided-depth-upsampling-using-anisotropic-total-generalized-variation:ICCV-2013} utilizes an anisotropic diffusion tensor to solve the problem of DSR. What's more, a bimodal co-sparse analysis model generated from color images are utilized in ~\cite{kiechle:A-joint-intensity-and-depth-co-sparse-analysis-model-for-depth-map-super-resolution:ICCV2013} to generate an HR depth map from an LR depth map. Additionally,~\cite{Matsuo:depth-image-enchancement-using-local-tangent-plane-approximations:CVPR-2015} proposes to compute local tangent planes using HR color images in the process of DSR, since it can provide auxiliary information. Besides, the consistency information between color images and depth maps is used to generate HR depth maps in~\cite{Lu:Sparse-depth-super-resolution:CVPR-2015}.


\vspace{-2.5mm}
\subsubsection{DCNN based DSR} The success of DCNN in high-level computer vision tasks has been extended to DSR. Using SRCNN~\cite{Learning-a-deep-convolutional-network-for-image-super-resolution:ECCV-2014} as the mapping unit, ~\cite{song-Deep-depth-super-resolution:ACCV2016} proposes an effective DCNN based learning method to generate a HR depth map from an LR depth map. Meanwhile,~\cite{Riegler:ATGV-Net:ECCV2016} proposed a ATGV-Net which combines DCNN with total variations to generate HR depth maps. The total variations are expressed by layers with fixed parameters. Besides, a novel DCNN based method is proposed in~\cite{riegler:A-deep-primal-dual-netwok-for-guided-depth-super-resolution:BMVC2016}, which combines a DCNN with a non-local variational method. Note that corresponding HR color images and up-sampled LR depth maps are regarded as input to feed into the network in~\cite{riegler:A-deep-primal-dual-netwok-for-guided-depth-super-resolution:BMVC2016}\cite{Riegler:ATGV-Net:ECCV2016}\cite{song-Deep-depth-super-resolution:ACCV2016}. What's more, a multi-scale fusion strategy is utilized in~\cite{hui:Depth-map-super-resolution-by-deep-multi-scale-guidance:ECCV2016}, which uses a multi-scale guided convolutional network for DSR with and without the guidance of the color images. Besides,~\cite{song-Deeply-supervised-depth-map-super-resolution:TCSVT2019} proposes a novel framework by using view synthesis to explain the generation of LR depth maps.~\cite{Voynov:perceptual-deep-depthsr:2019_ICCV} used rendering of 3D surfaces to measure the quality of obtained depth maps. It demonstrates that a simple visual appearance based loss yields significantly improved 3D shapes.

However, most of conventional learning based DSR and DCNN based DSR exploit bi-cubic degradation to generate LR depth maps, which are not enough to describe the generation of LR depth maps in real world.
\section{Our approach}

\begin{figure*}
  \centering
  \includegraphics[width=0.9\linewidth]{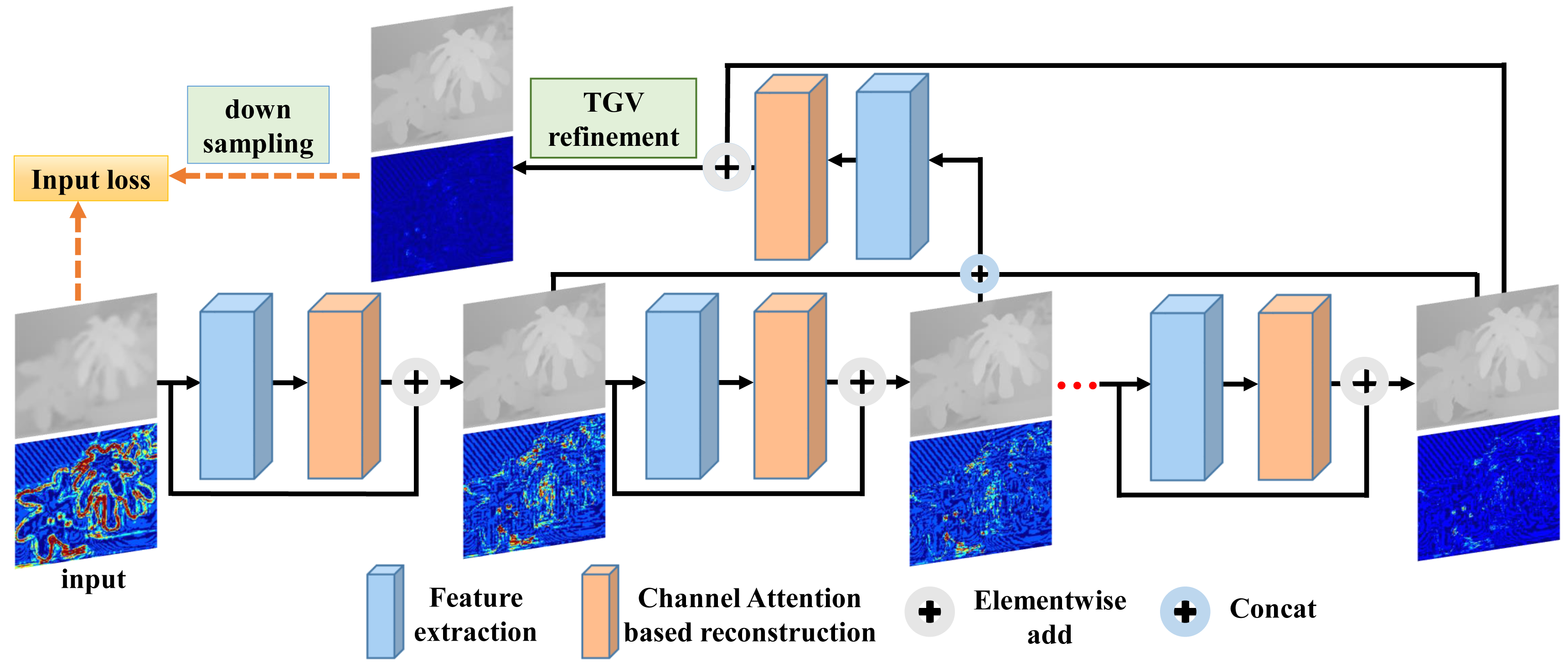}
  \caption{The figure shows the pipeline of the proposed framework. We show the residual of the output between each sub-module and groundtruth. From blue to red means value from 0 to $\infty$.}
  \label{fig:pipeline}
  \vspace{-0.4cm}
\end{figure*}

\subsection{Overview}

To tackle with different types of LR depth maps in DSR, including non-linear degradation with noise and interval down-sampling degradation, we adopt an iterative residual learning framework. As shown in Fig.~\ref{fig:pipeline}, it contains several sub-modules, and the input of each sub-module is the output the previous sub-module, which guarantees that the residual between the input and groundtruth can be learned step by step. Besides, as the network goes deeper, strong supervision from groundtruth is added in each sub-module to release gradient vanishing. In the last sub-module, high-frequency components obtained by previous sub-modules are fused as input, which re-exploits cause-to-fine high-level information to further improve the performance of the proposed framework. In each sub-module, residual learning strategy is used, and it contains two indispensable blocks: feature extraction block and channel attention based reconstruction block. Besides, except the loss between output and groundtruth, if it is well recovered, the down-sampled version of obtained depth maps should be same with the input $\bD^L$, hence, we use such input loss to constrain the framework. What's more, to maintain sharp boundaries, total generalized variation ($\mathsf{TGV}$) term is utilized to further refine the obtained HR depth maps.

\subsection{Network structure}

As shown in Fig.~\ref{fig:pipeline}, our network can be unfolded to $K$ sub-modules. We utilize a residual connection for sub-modules and calculate loss between each sub-module's output and groundtruth to alleviate gradient vanishing.
The loss function is defined in Sec.~\ref{subsec:loss}. Each sub-module contains two parts: feature extraction block (FE) and channel attention based reconstruction block (CAR). 


For the $k$-th ($k \in [1,K]$) sub-module, its input and output is defined  as $\bI^{k}$ and $\bO^k$, respectively. The operation of learning high-frequency component $\bO^{k}_{\text{CAR}}$ is given by:
\begin{equation}\label{Eq::FE_CAR}
    \begin{split}
        \bO^{k}_{\text{FE}} &= \sF_{\text{FE}}\left(\bI^{k}\right) \\
        \bO^{k}_{\text{CAR}} &= \sF_{\text{CAR}}\left(\bO^{k}_{\text{FE}}\right),
    \end{split}
\end{equation}
where $\sF_{\text{FE}}(\cdot)$ and $\sF_{\text{CAR}}(\cdot)$ are feature extraction and channel attention based reconstruction operation, respectively.

The output of $k$-th sub-module $\bO^k$ is given by:
\begin{equation}\label{Eq::add}
    \begin{split}
    \bO^k &= \bO^{k}_{\text{CAR}} + \bI^{k}.
    \end{split}
\end{equation}

Combing Eq.~\eqref{Eq::FE_CAR} and \eqref{Eq::add}, the operation of $k$-th sub-module can be summarized as:
\begin{equation}
    \bO^{k} = \sS_k\left(\bI^{k}\right),
\end{equation}
where $\sS_k(\cdot)$ denotes the operation of $k$-th sub-module. The output of $k$-th sub-module is taken as the input of the next sub-module, \ie $\bO^{k} = \bI^{k+1}$.

For the last sub-module $K$, the input is the concatenation of $\bO^1$ to $\bO^{K-1}$, dubbed $\sF_\text{concat}\left(\bO^1,\cdot\cdot\cdot, \bO^{K-1}\right)$, where $\sF_\text{concat}(\cdot)$ is the concatenation operation, and the operation of last sub-module $K$ is given by:

    

\begin{equation}
    \bO^{k} = \sS_k\left(\sF_\text{concat}\left(\bO^1, \cdot\cdot\cdot, \bO^{K-1}\right)\right).
\end{equation}

For the first sub-module, the input is the up-sampled version of LR depth maps $\bD^L$ ($\uparrow\lambda$ times, where $\lambda$ is the up-sampling factor). We use bi-cubic up-sample kernel for simplicity.

\subsection{Feature extraction block}
A convolutional layer is dubbed as $\text{Conv}\left(m,n\right)$, where $m$ is the kernel size and $n$ is the number of kernels. In feature extraction block, it contains $l$ convolutional layers with ReLU as activation function. We set $m=3$, $n = 64$ and $l = 8$ in this paper.  

\begin{figure}
    \centering
    \includegraphics[width=1.05\linewidth]{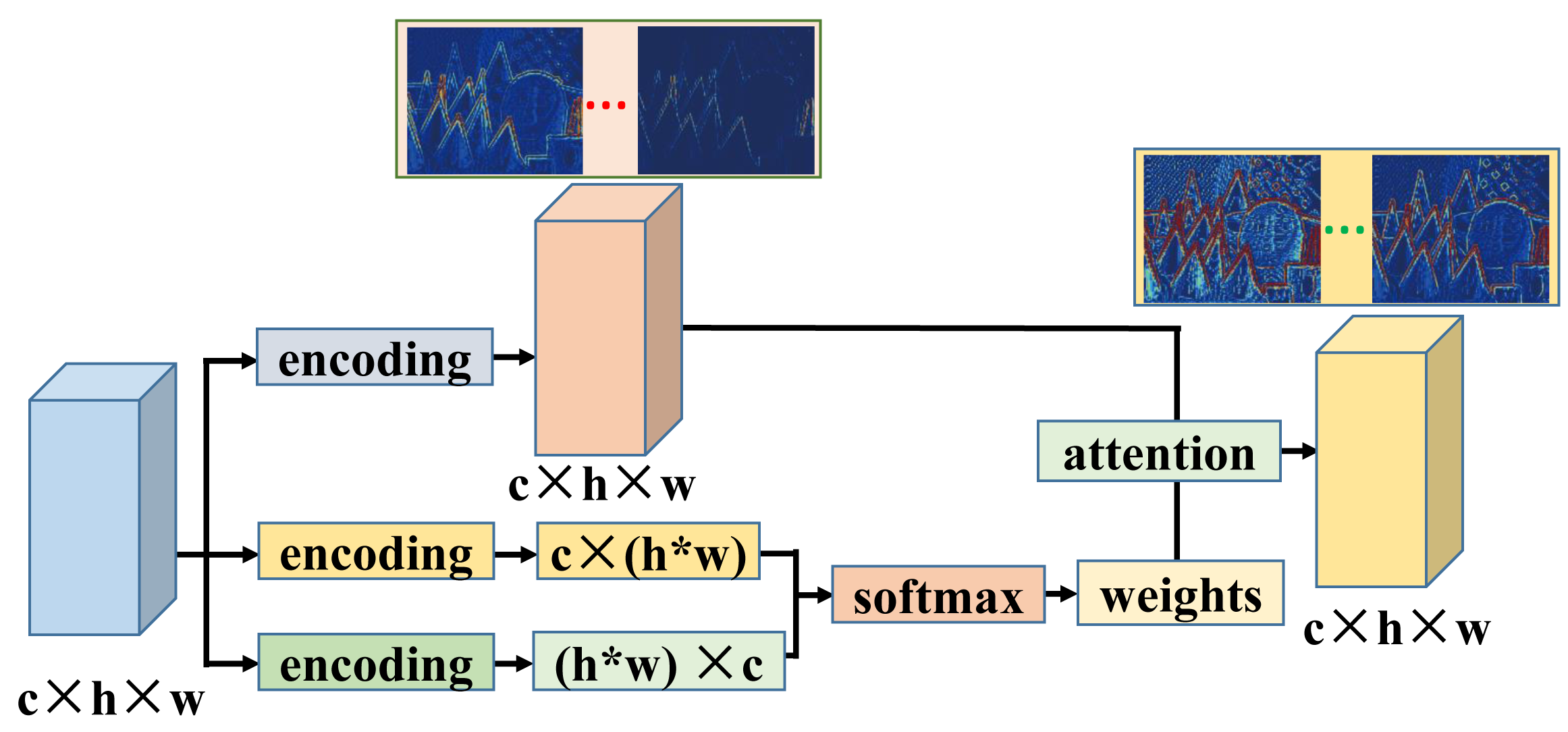}
    \caption{The figure shows the pipeline of channel attention.}
    \label{fig:channel_attention}
    \vspace{-0.6cm}
\end{figure}

\subsection{Channel attention based reconstruction}
Inspired by~\cite{Liu:ABPN-ICCVW2019}, we proposed to use attention strategy in DSR, and the proposed channel attention based reconstruction block provides imperative information to learn high-frequency components of depth maps. It contains two steps: channel attention and reconstruction.



\textbf{Channel attention:} Each sub-module $\sF_{\text{CAR}}(\cdot)$ takes the output of feature extraction block $\bO^{k}_\text{FE}$ as input. For $\bO^{k}_\text{FE}$  with tensor size of $\left(c\times h \times w\right)$, $\sF_{\text{CAR}}(\cdot)$ first converts $\bO^{k}_\text{FE}$ to three components 
$\sP\left(\bO^{k}_\text{FE}\right)$, $\sQ\left(\bO^{k}_\text{FE}\right)$ and $\sV\left(\bO^{k}_\text{FE}\right)$ via  encoding operations $\sP\left(\cdot\right)$, $\sQ\left(\cdot\right)$ and $\sV\left(\cdot\right)$, respectively. The tensor size of $\sP\left(\bO^{k}_\text{FE}\right)$, $\sQ\left(\bO^{k}_\text{FE}\right)$ and $\sV\left(\bO^{k}_\text{FE}\right)$ is $\left(c\times h \times w\right)$,  $\left(c\times hw\right)$ and $\left(hw \times c\right)$, respectively.


$\sP(\cdot)$ is data pre-processing and it contains $\alpha$ convolutional layer.
$\sQ(\cdot)$ and $\sV(\cdot)$ are convolution with reshape operations. The number of convolutional layers are $\beta$ and $\gamma$, respectively.
$\sQ(\cdot)$ and $\sV(\cdot)$ are defined for learning channel attention parameters.
$\sQ\left(\bO^{k}_\text{FE}\right)$ and $\sV\left(\bO^{k}_\text{FE}\right)$ are dot-producted (elementwise multiplication), and fed to a $\text{softmax}$ operation to regress channel attention weights $\btheta$.
$\sP\left(\bO^{k}_\text{FE}\right)$ and $\btheta$ are dot-producted to obtain the output of channel attention $\bO^{k}_{\text{CAR}}$.
Fig.~\ref{fig:channel_attention} shows the pipeline of the proposed channel attention. The above operations can be defined as following:
\begin{equation}
\begin{split}
    &\btheta = \text{softmax}\left(\sQ\left(\bO^{k}_\text{FE}\right) \odot \sV\left(\bO^{k}_\text{FE}\right)\transpose\right), \\
    &\bO^k_\text{CAR} = \btheta \odot \sP\left(\bO^{k}_\text{FE}\right).
\end{split}
\end{equation}

The channel attention can be understood as non-local convolution process, which aims to enhance the channels with much more effective information. The non-local operation in the proposed channel attention based reconstruction can obtain effective attention weights for each channel by exploiting all the position information of the feature maps. $\sQ\left(\bO^{k}_\text{FE}\right) \odot \sV\left(\bO^{k}_\text{FE}\right)\transpose$ can be regarded as a form of covariance of the input data. It provides an effective score to describe the tendency of two feature maps at different channels. 

\textbf{Reconstruction:} Based on $\bO^k_\text{CAR}$, we can obtain its reconstruction result $\bO^k$ by using $\eta$ convolutional layers. In this paper, we set $\alpha=\beta=\gamma=\eta=1$, and use $\text{Conv}\left(3, 64\right)$ in channel attention stage and $\text{Conv}\left(3,1\right)$ in reconstruction stage. The effectiveness of the proposed channel attention based reconstruction block will be demonstrated in the experiment section.

\subsection{Loss Function} \label{subsec:loss}
We use the $L_1$ loss to optimize the proposed framework.

\subsubsection{Sub-module loss} 

For the $k$-th sub-module, the loss is defined as:
\begin{equation}
    L_k = {||\bO^k - \bD^G||}_{1},
\end{equation}
where $L_k$ is the loss between the output of $k$-th sub-module and the groundtruth. 

Our framework can obtain $K$ HR depth maps with LR depth maps as input. Generally, one will pay more attention on the output of the last sub-module, hence different weights are set for losses at different sub-modules, and the loss weight increases as the network goes deeper. The final loss for sub-modules is defined as following: 

\vspace{-0.2in}
\begin{equation}
\begin{split}
    L_s = 
    \sum_{k=1}^{K}\frac{k}{N}L_k,
\end{split}
\end{equation}

where $N= \sum_{k=1}^{K} k = K(K+1)/2$.

\vspace{-1.5mm}
\subsubsection{Input loss and TGV term}
The HR depth map is well recovered, the down-sampled version (same degradation model) of the finally obtained depth maps should be the same as the original LR input $\bD^L$. Hence, we use the input loss to further constrain the obtained HR depth map, which is defined as:
\begin{equation}
    L_{\text{input}} = {||\sF_{\text{down}}\left(\bO^{K}\right) - \bD^{L}||}_{1},
\end{equation}
where $\sF_{\text{down}}(\cdot)$ is the degradation model, with output tensor of the same size as $\bD^L$. 

Besides, depth maps usually contain sharp boundaries, hence, the total generalized variation $\mathsf{TGV}\left(\bO^{K}\right)$ is exploited to refine the final obtained HR depth maps.

The final loss of our proposed framework is defined as:
\vspace{-1.5mm}
\begin{equation}
    L = L_s + \xi_1 L_{\text{input}} + \xi_2 \mathsf{TGV}(\bO^{K}),
\end{equation}
where $\xi_1$ and $\xi_2$ are weights for input loss and total variation term. We set $\xi_1 = 0.1$ and $\xi_2 = 0.05$ in this paper.

\subsection{Implementation}

We employed Adam~\cite{kingma2014adam} as the optimizer to optimize the parameters, and the learning rate varies from $0.1$ to $0.0001$ by multiplying 0.1 for every 25 epochs. Adjustable gradient clipping strategy~\cite{Kim:Accurate-image-super-resolution-using-very-deep-convolutional-networks:CVPR2016} is used. The proposed framework converged after 100 epochs.
\section{Experiment}

In this section, we evaluate the performance our method against different state-of-the-art (SOTA) methods on diverse publicly available datasets using different types of LR depth maps as input, including non-linear degradation with noise and interval down-sampling degradation. 

\subsection{Datasets}


We used 6 different datasets in this paper: (1). Middlebury dataset~\cite{Evaluation-of-cost-functions-for-stereo-matching:CVPR-2007}\cite{middlebury-2014:GCPR2014}\cite{Learning-conditional-random-fields-for-stereo:CVPR-2007}\cite{A-taxonomy-and-evaluation-of-dense-two-frame-stereo-correspondence-algorithms:IJCV-2002}, which provides high-quality depth maps for complex real-world scenes; (2). The Laserscan dataset~\cite{Patch-based-synthesis-for-single-depth-image-super-resolution:ECCV-2012}, which is captured by laser sensors and provides accurate depth measurements; (3). Sintel dataset~\cite{Butler:A-Naturalistic-open-source-movie-for-optical-flow-evaluation:ECCV:2012}, ICL dataset~\cite{ICLdataset:ICRA2014} and synthetic New Tsukuba dataset~\cite{Martull:Realistic-CG-stereo-image-dataset-with-ground-truth-disparity-maps:ICPR2012}, which are synthesized datasets and contain lots of depth details and high quality depth maps; (4). SUN RGBD dataset~\cite{sug-rgbd-dataset:CVPR2015}, which contains images captured with different consumer-level RGBD cameras, such as Microsoft Kinect, and provides low-quality depth maps for complex real-world scenes; and (5). Apolloscape dataset~\cite{apolloscape_1}\cite{apolloscape_2}, which contains high-quality depth maps captured by Lidar in real traffic scenes.

\textbf{Training dataset:} To effectively training the proposed framework, we follow  DVS~\cite{song-Deeply-supervised-depth-map-super-resolution:TCSVT2019} to prepare our training dataset. $115$ depth maps are collected from the Middlebury dataset~\cite{Evaluation-of-cost-functions-for-stereo-matching:CVPR-2007}\cite{Learning-conditional-random-fields-for-stereo:CVPR-2007}\cite{A-taxonomy-and-evaluation-of-dense-two-frame-stereo-correspondence-algorithms:IJCV-2002}, the Sintel dataset~\cite{Butler:A-Naturalistic-open-source-movie-for-optical-flow-evaluation:ECCV:2012} and the synthetic New Tsukuba dataset~\cite{Martull:Realistic-CG-stereo-image-dataset-with-ground-truth-disparity-maps:ICPR2012}. Using these depth maps, the input LR depth maps $\mathbf{D}^{L}$ are obtained by $\mathbf{D}^{L} = \downarrow_\lambda \mathbf{D}^{G}$, where $\lambda$ is the down-sampling factor. To simulate the generation of real LR depth maps, non-linear degradation with noise and interval down-sampling degradation are used. Besides,  bi-cubic downsampling is commonly used in DSR~\cite{song-Deep-depth-super-resolution:ACCV2016}\cite{song-Deeply-supervised-depth-map-super-resolution:TCSVT2019}\cite{Xie:edge-guided-single-depth-sr:TIP2015}, hence, we use bi-cubic degradation to evaluate the performance of non-linear degradation.



\begin{figure}
  \centering
  \includegraphics[width=0.8\linewidth]{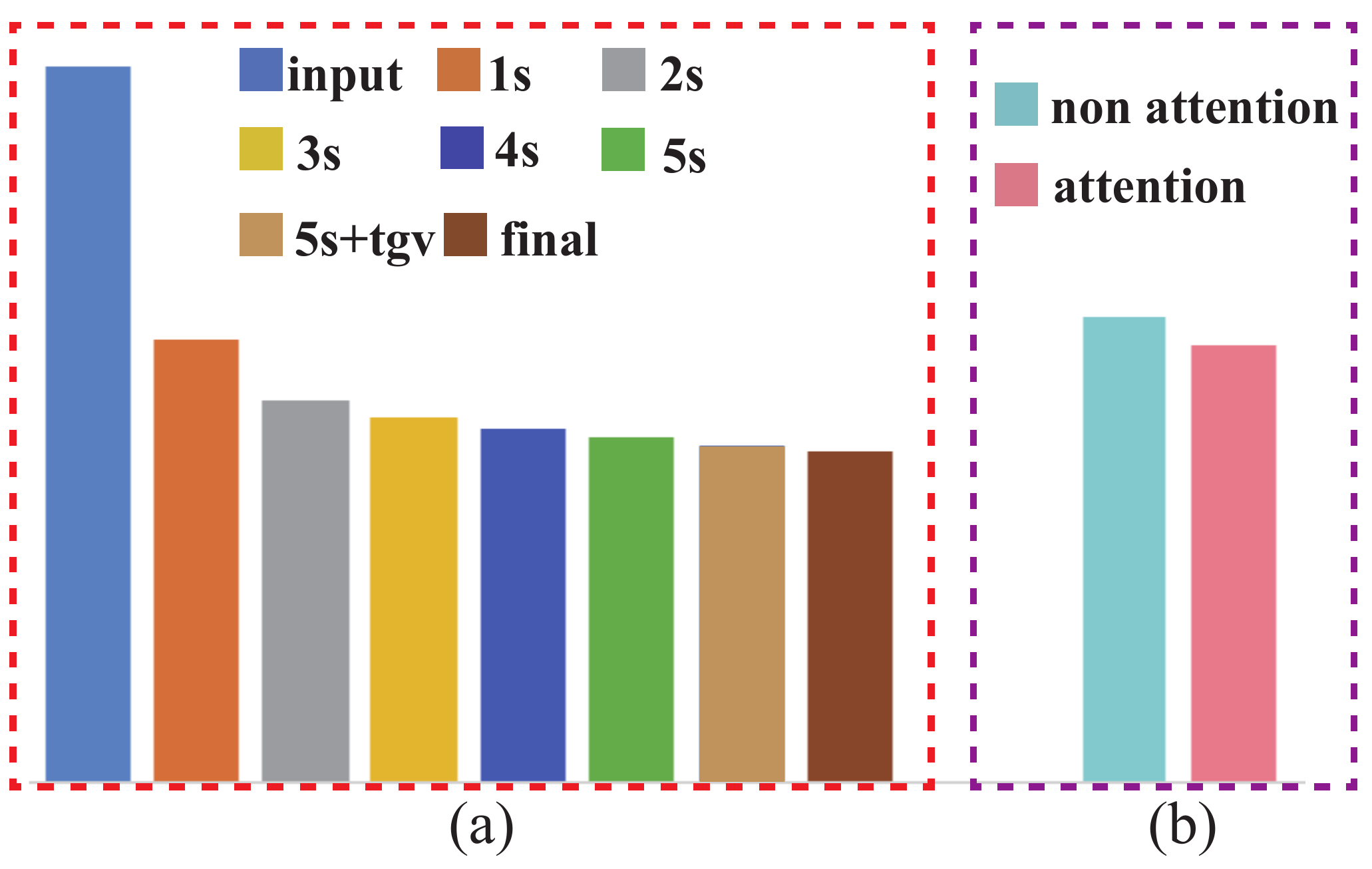}
  \caption{(a) shows the results of average $RMSE$ of the proposed framework with different number of sub-modules. $1s$ to $5s$ are number of sub-modules from 1 to 5 respectively. $5s+tgv$ means 5 sub-modules with TGV refinement and $final$ means 5 sub-modules with input loss and TGV refinement. (b) shows the average $RMSE$ results of attention and non-attention respectively.} 
  \label{fig:ablation_study&attention}
  \vspace{-0.4cm}
\end{figure}

\begin{figure}
  \centering
  \includegraphics[width=0.9\linewidth]{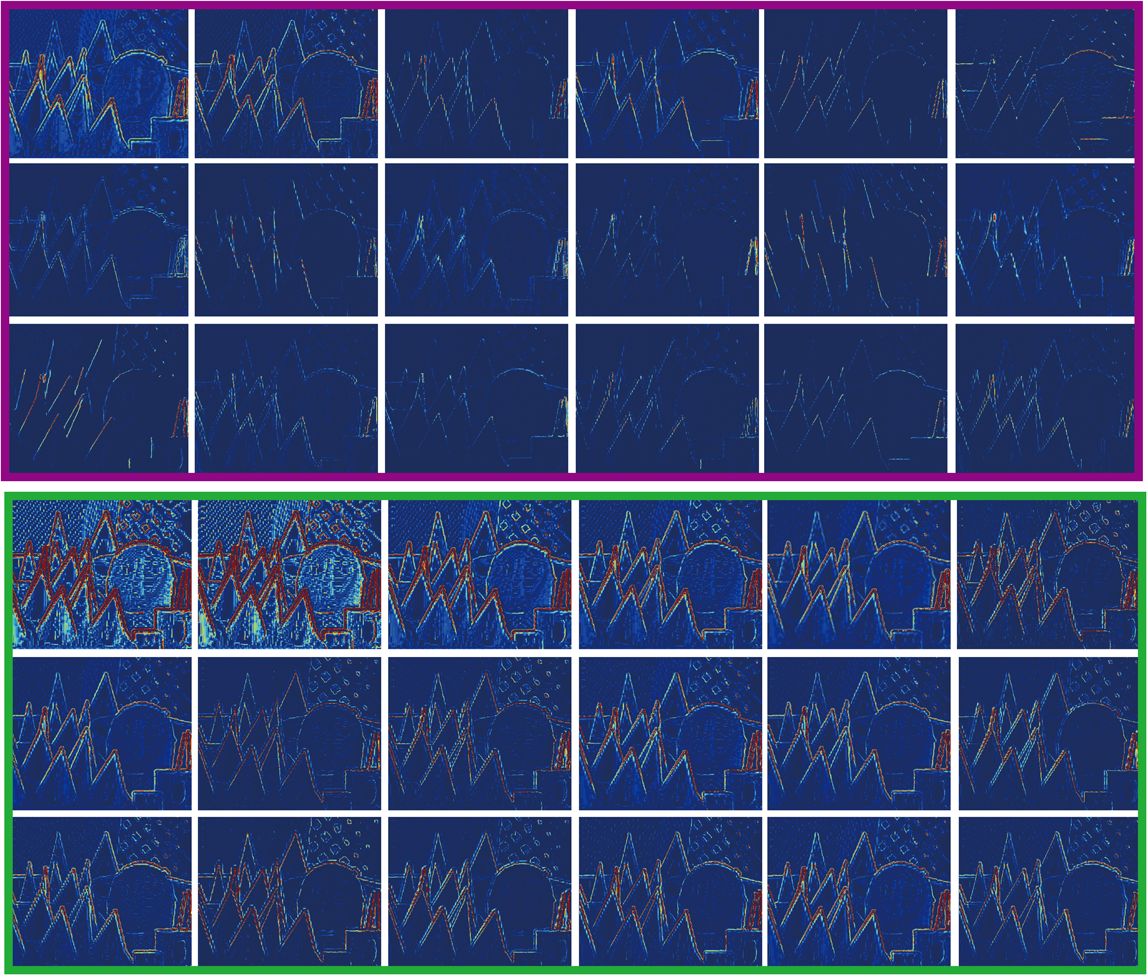}
  \caption{The feature maps of high-frequency component before and after channel attention. Purple and green areas shows the feature maps before and after channel attention block respectively. From blue to red means value from to 0 to $\infty$. Best viewed on screen.} 
  \label{fig:useful_attention}
  \vspace{-0.5cm}
\end{figure}

\begin{figure*}
  \centering
  \includegraphics[width=0.96\linewidth]{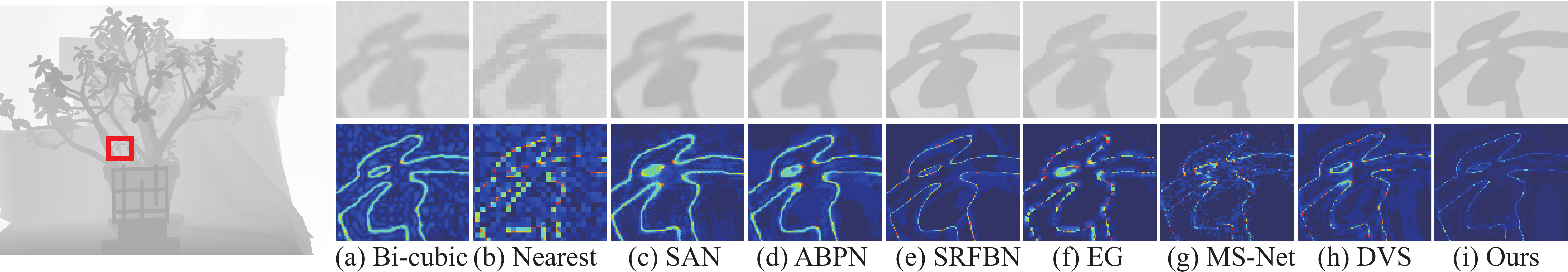}
  \caption{Comparison on Middlebury 2014 dataset~\cite{middlebury-2014:GCPR2014} ($Jadeplant$) under up-upsampling factor of $\times 4$ (bi-cubic degradation with noise). (a) Bi-cubic, (b) Nearest Neighbor, (c) SAN~\cite{Dai_second-order-sr-2019_CVPR}, (d) ABPN~\cite{Liu:ABPN-ICCVW2019}, (e) SRFBN~\cite{Li-SRFBN:CVPR2019}, (f) EG~\cite{Xie:edge-guided-single-depth-sr:TIP2015}, (g) MS-Net~\cite{hui:Depth-map-super-resolution-by-deep-multi-scale-guidance:ECCV2016}, (h) DVS~\cite{song-Deeply-supervised-depth-map-super-resolution:TCSVT2019} and (i) Our results. The second row shows the residual between the results and groundtruth. From blue to red means 0 to $\infty$. Best viewed on screen.} 
  \label{fig:results_noise_x4}
  \vspace{-0.3cm}
\end{figure*}

\begin{figure*}
  \centering
  \includegraphics[width=0.96\linewidth]{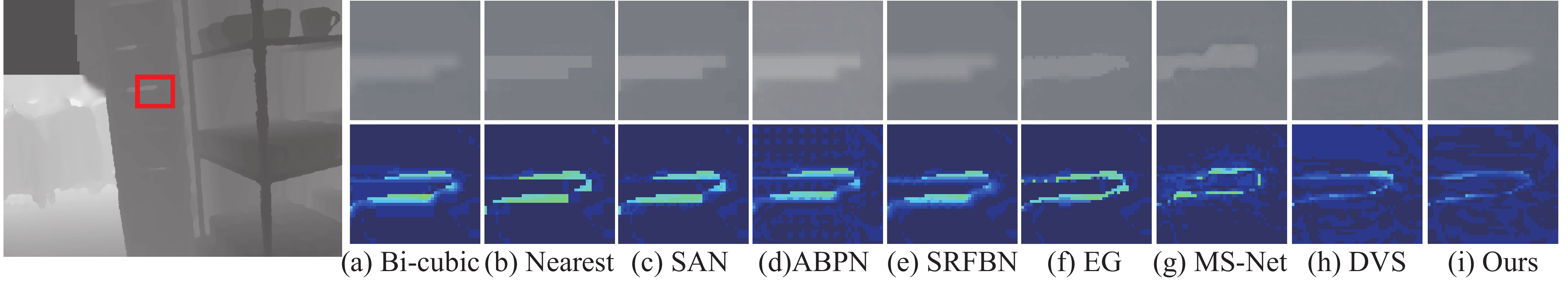}
  \vspace{-0.3cm}
  \caption{Comparison on depth map captured by Kinect~\cite{sug-rgbd-dataset:CVPR2015} ($0100$) under up-upsampling factor of $\times 4$ (interval down-sampling degradation) on depth map captured by Kinect. (a) Bi-cubic, (b) Nearest Neighbor, (c) SAN~\cite{Dai_second-order-sr-2019_CVPR}, (d) ABPN~\cite{Liu:ABPN-ICCVW2019}, (e) SRFBN~\cite{Li-SRFBN:CVPR2019}, (f) EG~\cite{Xie:edge-guided-single-depth-sr:TIP2015}, (g) MS-Net~\cite{hui:Depth-map-super-resolution-by-deep-multi-scale-guidance:ECCV2016}, (h) DVS~\cite{song-Deeply-supervised-depth-map-super-resolution:TCSVT2019} and (i) Our results. The second row shows the residual between the results and groundtruth. From blue to red means 0 to $\infty$. Best viewed on screen.} 
  \label{fig:results_int_x4}
  \vspace{-0.4cm}
\end{figure*}

\begin{figure*}
  \centering
  \includegraphics[width=0.96\linewidth]{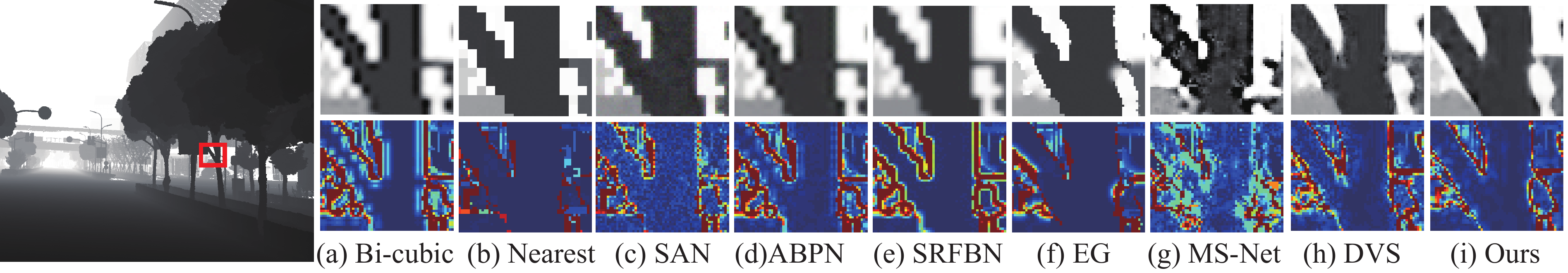}
  \vspace{-1.5mm}
  \caption{Comparison on Lidar data ($road01$ of Apolloscape dataset~\cite{apolloscape_1}\cite{apolloscape_2}) in real traffic scenes under up-upsampling factor of $\times 4$ (interval down-sampling degradation). (a) Bi-cubic, (b) Nearest Neighbor, (c) SAN~\cite{Dai_second-order-sr-2019_CVPR}, (d) ABPN~\cite{Liu:ABPN-ICCVW2019}, (e) SRFBN~\cite{Li-SRFBN:CVPR2019}, (f) EG~\cite{Xie:edge-guided-single-depth-sr:TIP2015}, (g) MS-Net~\cite{hui:Depth-map-super-resolution-by-deep-multi-scale-guidance:ECCV2016}, (h) DVS~\cite{song-Deeply-supervised-depth-map-super-resolution:TCSVT2019} and (i) Our results. The second row shows the residual between the results and groundtruth. From blue to red means 0 to $\infty$. Best viewed on screen.} 
  \label{fig:results_lidar_int_x4}
  \vspace{-0.4cm}
\end{figure*}

\begin{table*}[]
    \centering
    \footnotesize
    \begin{tabular}{c c c c c c c c c c c c }  \hline
        $\times{4}$ & $Plant$ & $Room$ & $0100$ & $0400$ & $Motorcycle$ & $Playtable$ & $Flowers$ & $Jadeplant$ & $ls21$ & $ls30$ & $ls42$ \\ \hline
        Bi-cubic  & 1.2340 & 1.5448 & 3.0922 & 1.3039 & 4.9046 & 2.9967 & 4.6655 & 4.1660 & 2.8441 & 2.6544 & 5.4735 \\ \hline
        Nearest & 1.4102 & 1.7558 & 3.4003 & 1.5271 & 5.6645 & 3.4443 & 5.4189 & 4.8238 & 3.3306 & 3.1039 & 6.3581 \\ \hline
        ABPN~\cite{Liu:ABPN-ICCVW2019} & 1.1588 & 1.2605 & 2.8357 & 1.1167 & 4.6597 & 2.7904 & 4.4472 & 4.0635 & 2.5961 & 2.5063 & 5.1318 \\ \hline
        SRFBN~\cite{Li-SRFBN:CVPR2019} & 1.1039 & 1.3029 & 2.8254 & 1.0808 & 4.3934  & 2.5663 & 4.0677 & 3.6864 & 2.4516 & 2.2565 & 4.9645 \\ \hline
        SAN~\cite{Dai_second-order-sr-2019_CVPR} & 1.2297 & 1.3813 & 2.9248 & 1.1987 & 4.4938 & 2.6558 & 4.2388 & 3.9288 & 2.5027 & 2.3519 & 5.0777 \\ \hline
        IKC~\cite{Gu_blendsr:2019_CVPR} & 1.2048 & 1.3240 & 2.9011 & 1.1324 & 4.4215 & 2.6078 & 4.1846 & 3.8026 & 2.4865 & 2.3028 & 4.9981 \\ \hline
        EG~\cite{Xie:edge-guided-single-depth-sr:TIP2015} & 1.4253 & 1.7250 & 3.3987 & 1.5038 & 5.4685 & 3.3261 & 5.2067 & 4.6162 & 3.2764 & 6.0576 & 6.4288 \\ \hline
        MS-Net~\cite{hui:Depth-map-super-resolution-by-deep-multi-scale-guidance:ECCV2016} & 1.1952 & 1.5116 & 3.1302 & 3.6576 & 5.0119 & 2.9683 & 4.7982 & 4.2426 & 2.7356 & 2.6127 & 5.8623 \\ \hline
        DVS~\cite{song-Deeply-supervised-depth-map-super-resolution:TCSVT2019} & 0.8494 & 1.1682 & 2.8914 & 0.9601 & 3.2553 & 2.0168 & 3.0409 & 2.9407 & 1.8188 & 1.8079 & 3.2001 \\ \hline
        $AIR_{ws}(ours)$ & \underline{0.7300} & \underline{1.0952} & \underline{2.8028} & \underline{0.7531} & \underline{3.1025} & \underline{1.9024} & \underline{2.9520} & \underline{2.8004} & \underline{1.6252} & \underline{1.5930} & \underline{2.9528} \\ \hline
        $AIR(ours)$ & \textbf{0.7278} & \textbf{1.0639} & \textbf{2.7800} & \textbf{0.7611} & \textbf{3.0968} & \textbf{1.8626} & \textbf{2.8873} & \textbf{2.7740} & \textbf{1.6048} & \textbf{1.5668} & \textbf{2.9332} \\ \hline
    \end{tabular}
    \caption{Comparison of $RMSE$ results under up-sampling factor of $\times{4}$ (interval down-sampling degradation). $AIR_{ws}$ means results obtained by weights sharing among different sub-modules and $AIR$ means non weights sharing. The best result is highlighted and the second best is underlined.}
    \label{tab:rmse_x4_int}
    \vspace{-0.3cm}
\end{table*}

\textbf{Evaluation:} To effectively evaluate the performance of our proposed framework, depth maps ($Motorcycle$, $Playtable$, $Flowers$ and $Jadeplant$) from Middlebury 2014 dataset~\cite{middlebury-2014:GCPR2014} are chosen as the testing depth maps. Besides, to further evaluate the generalization performance of the proposed framework, we also evaluate depth maps chosen from ICL dataset~\cite{ICLdataset:ICRA2014} ($Plant$ and $Room$), Laser-Scan dataset~\cite{Patch-based-synthesis-for-single-depth-image-super-resolution:ECCV-2012} ($ls21$, $ls30$ and $ls42$), SUN RGBD dataset~\cite{sug-rgbd-dataset:CVPR2015} ($0100$ and $0400$) and Apolloscape dataset~\cite{apolloscape_1}\cite{apolloscape_2}($road01$, $road05$, $road10$ and $road 17$). Note that models are trained with depth maps from Middlebury dataset, sintel dataset and synthetic New Tsukuba dataset.

\begin{table*}[]
    \centering
    \footnotesize
    \begin{tabular}{c c c c c c c c c c c c }  \hline
        $\times{4}$ & $Plant$ & $Room$ & $0100$ & $0400$ & $Motorcycle$ & $Playtable$ & $Flowers$ & $Jadeplant$ & $ls21$ & $ls30$ & $ls42$ \\ \hline
        Bi-cubic  & 0.7300 & 0.9613 & 2.2028 & 0.8189 & 3.0434 & 1.8108 & 2.9092 & 2.6154 & 1.6278 & 1.5801 & 3.3351 \\ \hline
        Nearest & 0.8841 & 1.1528 & 2.3911 & 1.0324 & 3.7067 & 2.2046 & 3.5688 & 3.1795 & 2.1406 & 2.0200 & 4.2166 \\ \hline
        ABPN~\cite{Liu:ABPN-ICCVW2019} & 0.6561 & 0.8476 & 1.5899 & 0.8005 & 2.0048 & 1.2335 & 1.7998 & 1.7204 & 1.3236 & 1.1814 & 1.7379 \\ \hline
        SRFBN~\cite{Li-SRFBN:CVPR2019} & 0.7068 & 0.8727 & 1.4063 & 0.7885 & 1.8300 & 1.1699 & 1.6949 & 1.6797 & 1.2107 & 1.1710 & 1.6175 \\ \hline
        SAN~\cite{Dai_second-order-sr-2019_CVPR} & 0.6651 & 0.7238 & 1.7139 & 0.7588 & 2.0501 & 1.4477 & 1.9034 & 1.8425 & 1.4803 & 1.3692 & 1.8232 \\ \hline
        Meta-SR~\cite{Hu_Meta-SR:2019_CVPR} & 0.6467 & 0.8026 & 1.5319 & 0.8005 & 1.9938 & 1.2517 & 1.7581 & 1.7065 & 1.3016 & 1.1645 & 1.7158 \\ \hline
        IKC~\cite{Gu_blendsr:2019_CVPR} & 0.6815 & 0.7523 & 1.4652 & 0.7630 & 2.0812 & 1.3420 & 1.8351 & 1.8092 & 1.3521 & 1.2021 & 1.7956 \\ \hline
        EG~\cite{Xie:edge-guided-single-depth-sr:TIP2015} & 0.7740 & 0.9972 & 2.1337 & 0.8874 & 2.9183 & 1.6414 & 2.6186 & 2.5365 & 1.7593 & 1.6318 & 3.3086 \\ \hline
        MS-Net~\cite{hui:Depth-map-super-resolution-by-deep-multi-scale-guidance:ECCV2016} & 0.4675 & 0.6453 & 1.0524 & 0.5760 & 2.0554 & 1.3518 & 1.9564 & 1.9218 & 1.4324 & 1.4087 & 1.7569 \\ \hline
        DVS~\cite{song-Deeply-supervised-depth-map-super-resolution:TCSVT2019} & 0.4565 & 0.5903 & 0.9826 & 0.5387 & 1.9718 & 1.2588 & 1.8532 & 1.8458 & 1.3800 & 1.3424 & 1.7212 \\ \hline
        
        $AIR_{ws}(ours)$ & 0.4101 & \textbf{0.5196} & \underline{0.9692} & \underline{0.4996} & \underline{1.7923} & \underline{1.1655} & \underline{1.7324} & \underline{1.6781} & \underline{1.1456} & \underline{1.0788} & \textbf{1.4875} \\ \hline
        $AIR(ours)$ & \textbf{0.4004} & \underline{0.5351} & \textbf{0.9588} & \textbf{0.4986} & \textbf{1.7764} & \textbf{1.1622} & \textbf{1.7005} & \textbf{1.6765} & \textbf{1.1393} & \textbf{1.0633} & \underline{1.4877} \\ \hline
    \end{tabular}
    \caption{Comparison of the $RMSE$ results under up-sampling factor of $\times{4}$ (bi-cubic degradation with noise). $AIR_{ws}$ means weights sharing among different sub-modules and $AIR$ means non weights sharing. The best result is highlighted and the second best is underlined.}
    \label{tab:rmse_x4_noise_bic}
     \vspace{-0.3cm}
\end{table*}

\begin{table}[]
    \centering
    \footnotesize
    \begin{tabular}{ccccc}  \hline
        $\times{4}$ & $road01$ & $road05$& $road10$ & $road 17$ \\ \hline
        Bi-cubic  & 18.5311 & 33.6010 & 20.5177 & 19.1795  \\ \hline
        Nearest & 21.2863 & 38.6045 & 23.4327 & 22.1853  \\ \hline
        ABPN~\cite{Liu:ABPN-ICCVW2019} & 16.9054 & 28.4027 & 18.2029 & 17.3051 \\ \hline
        SRFBN~\cite{Li-SRFBN:CVPR2019} & 15.9080 & 27.9377 & 17.1010 & 16.0810  \\ \hline
        SAN~\cite{Dai_second-order-sr-2019_CVPR} & 16.1057 & 29.2059 & 18.5678 & 17.2564 \\ \hline
        IKC~\cite{Gu_blendsr:2019_CVPR} & 16.0557 & 28.3142 & 17.5034 & 16.4750 \\ \hline
        EG~\cite{Xie:edge-guided-single-depth-sr:TIP2015} & 27.7714 & 49.5420 & 31.0133 & 34.4618 \\ \hline
        MS-Net~\cite{hui:Depth-map-super-resolution-by-deep-multi-scale-guidance:ECCV2016} & 19.0029 & 31.8750 & 21.1448 & 20.1059 \\ \hline
        DVS~\cite{song-Deeply-supervised-depth-map-super-resolution:TCSVT2019} & 16.0110 & 26.4482 & 17.0613 & 16.0500 \\ \hline
        $AIR_{ws}(ours)$ & \underline{15.6305} & \underline{25.9282} & \underline{16.6152} & \underline{15.8423} \\ \hline
        $AIR(ours)$ & \textbf{15.6239} & \textbf{25.9109} & \textbf{16.5906} & \textbf{15.7792} \\ \hline
    \end{tabular}
    \caption{Comparison of $RMSE$ results under up-sampling factor of $\times{4}$ (interval down-sampling degradation). $AIR_{ws}$ means weights sharing among different sub-modules and $AIR$ means non weights sharing. The best result is highlighted and the second best is underlined.}
    \label{tab:rmse_x4_int_lidar}
    \vspace{-0.5cm}
\end{table}

\textbf{Baseline Methods:} Our propose method is compared with the following three categories of methods: (1). Standard interpolation approaches: Bi-cubic and Nearest Neighbour ($Nearest$); (2). State-of-the-art CNN based DSR approaches: EG~\cite{Xie:edge-guided-single-depth-sr:TIP2015}, MS-Net~\cite{hui:Depth-map-super-resolution-by-deep-multi-scale-guidance:ECCV2016}, DVS \etal~\cite{song-Deeply-supervised-depth-map-super-resolution:TCSVT2019}; (3). State-of-the-art CNN based color image super-resolution approaches: SRFBN~\cite{Li-SRFBN:CVPR2019}, Meta-SR~\cite{Hu_Meta-SR:2019_CVPR}, SAN~\cite{Dai_second-order-sr-2019_CVPR}, ABPN~\cite{Liu:ABPN-ICCVW2019} and IKC~\cite{Gu_blendsr:2019_CVPR}. Besides, all the methods are retrained with the same depth maps.

\textbf{Error metrics:} Root Mean Squared Error ($RMSE$) is used to evaluate the performance obtained by our method and other state-of-the-art methods. Specifically, $RMSE = \sqrt{\sum_{i=1}^N{(O_{i} - D^G_{i})^2}/N}$, where $O$ and $D^{G}$ are the obtained HR depth map and ground truth respectively, $N$ is the number of pixels in the HR depth map.
\vspace{-1mm}
\subsection{Ablation analysis}
\vspace{-1mm}
Fig.~\ref{fig:ablation_study&attention} (a) demonstrates the average $RMSE$ results of the proposed method with different number of sub-modules on Middlebury dataset ($Cones$, $Teddy$, $Tsukuba$ and $Venus$) under up-sampling factor of $\times 4$ (bi-cubic degradation). It can be observed that the $RMSE$ loss drops with the number of sub-module increases, which proves that deeper network with more sub-modules can obtain the residual component effectively, from which high-frequency component can be recovered step by step. Generally, any number of sub-module can be used in the proposed framework, and as shown in Fig.~\ref{fig:ablation_study&attention} (a), as the number of sub-module increasing, the whole framework becomes convergent. Hence, we set the number of sub-module $K=5$ in this paper. What's more, HR depth maps obtained by input loss and $TGV$ refinement get smaller $RMSE$, which demonstrates that these operations can recover more useful high-frequency information and further refine the obtained HR depth maps. Therefore, we can conclude that the all the components utilized in our framework contribute positively toward the final success of our approach.

\vspace{-1mm}
\subsection{Attention analysis}
\vspace{-1mm}
Fig.~\ref{fig:useful_attention} shows the feature maps obtained before and after channel attention strategy. The top $18$ feature maps with high-frequency component are shown in Fig.~\ref{fig:useful_attention}, purple and green areas demonstrate the feature maps before and after channel attention, respectively. And from blue to red means value from $0$ to $\infty$. According to Fig.~\ref{fig:useful_attention}, we can see that effective high-frequency component, such as edges, are efficiently enhanced by channel attention, which can be utilized to reconstruct better HR depth maps. Fig.~\ref{fig:ablation_study&attention} (b) shows the average $RMSE$ results of the proposed method with and with channel attention strategy. Middlebury dataset ($cones$, $teddy$, $tsukuba$ and $venus$) under up-sampling factor of $\times 4$ are used as input. It is obviously to find that smaller $RMSE$ can be obtained using channel attention strategy, which proves that the proposed channel attention strategy works positively in super-resolving DSR problem. 

According to Fig.~\ref{fig:useful_attention} and Fig.~\ref{fig:ablation_study&attention} (b), we can conclude that channel attention can enhance the channels with useful high-frequency information and improve the ability of each sub-module to obtain the residual, thus, recover high quality depth maps effectively.

\vspace{-1mm}
\subsection{Interval degradation} \label{clean_input}
\vspace{-1mm}

We first evaluate the performance of the proposed approach on depth maps with interval down-sampling degradation. The quantitative results in terms of $RMSE$ of up-sampling factors of $\times 4$ are reported in Table.~\ref{tab:rmse_x4_int} and Table.~\ref{tab:rmse_x4_int_lidar}. As indicated in Table.~\ref{tab:rmse_x4_int} and Table.~\ref{tab:rmse_x4_int_lidar}, $AIR_{ws}$ and $AIR$ demonstrate the results of the proposed method with and without weights-sharing among different sub-modules, respectively. It can be observed that the performances of state-of-the-art on interval down-sampled LR depth maps are not good enough (both CSR and DSR methods), and the proposed method outperforms other DCNN based methods with smaller $RMSE$. Besides, Table.~\ref{tab:rmse_x4_int_lidar} shows the results on dense depth maps captured by Lidar on real traffic scenes, which proves that the proposed framework can tackle with real LR Lidar data effectively. Besides, we can see that results of weights-sharing outperforms other state-of-the-art methods, and results of non-weight-sharing obtain better $RMSE$ results because it contains more parameters, thus have stronger non-linear mapping abilities to recover better HR depth maps.

Qualitative results are illustrated in Fig.~\ref{fig:results_int_x4} ($0100$ extracted from SUN RGBD dataset~\cite{sug-rgbd-dataset:CVPR2015}) and Fig.~\ref{fig:results_lidar_int_x4} ($road01$ from Apolloscape dataset~\cite{apolloscape_1}\cite{apolloscape_2}) for an up-sampling factor $\times4$ under interval down-sampling degradation. As shown in Fig.~\ref{fig:results_int_x4} and Fig.~\ref{fig:results_lidar_int_x4}, $0100$ and $road01$ are depth maps captured by Kinect and Lidar, which represent the depth maps captured in indoor and outdoor scenes of real world. Obviously, the proposed method produces more visually appealing results with smaller residual compared with groundtruth. Boundaries generated by the proposed method are sharper and more accurate, which demonstrate that the structure and high-frequency component of high-resolution depth maps can be well recovered.

\vspace{-1mm}
\subsection{Bi-cubic degradation}\label{noisy_input}
\vspace{-1mm}
In this section, we evaluate the proposed framework on noisy depth maps. Following~\cite{Riegler:ATGV-Net:ECCV2016}\cite{song-Deeply-supervised-depth-map-super-resolution:TCSVT2019}, depth dependent Gaussian noise is added to LR depth maps $\mathbf{D}^{LR}$ in the form $\theta(d) = \mathcal{N}(0,\delta/d)$, where $\delta = 651$ and $d$ denotes the depth value of each pixel in $\mathbf{D}^{L}$. Besides, to evaluate the ability of noise handling, we also add noise on depth maps captured by Kinect ($0100$ and $0400$ from SUN RGBD dataset~\cite{sug-rgbd-dataset:CVPR2015}).

Table~\ref{tab:rmse_x4_noise_bic} reports the quantitative results in terms of $RMSE$ for the up-sampling factor of $\times 4$ with bi-cubic degradation and noise as input, from which, we can clearly see that the proposed method outperforms others, even on raw depth maps with additional added noise ($0100$ and $0400$). The proposed method can well eliminate the influence of noise, thus depth maps with smaller $RMSE$ can be obtained.

Fig.~\ref{fig:results_noise_x4} illustrates the qualitative results of the proposed method ($Jadeplant$ from Middlebury 2014 dataset~\cite{middlebury-2014:GCPR2014}) under up-sampling factor $\times 4$ with bi-cubic degradation and noise as input. As shown in Fig.~\ref{fig:results_noise_x4}, $Jadeplant$ contains complex textures and luxuriant details, which is hard to recover a HR depth map from a LR depth map. Obviously, the proposed method produces more visually appealing results with sharper and more accurate boundaries, which proves that the proposed method can effectively recover the structure of HR depth maps.

\vspace{-1mm}
\subsection{Generalization ability}
\vspace{-1mm}

As discussed in section~\ref{clean_input} and section~\ref{noisy_input}, depth maps of ICL dataset~\cite{ICLdataset:ICRA2014}, SUN RGBD dataset~\cite{sug-rgbd-dataset:CVPR2015}, Laserscan dataset~\cite{Patch-based-synthesis-for-single-depth-image-super-resolution:ECCV-2012} and Apolloscape dataset~\cite{apolloscape_1}\cite{apolloscape_2} are not included in the training data. Based on Table.~\ref{tab:rmse_x4_int}, Table.~\ref{tab:rmse_x4_noise_bic}, Table.~\ref{tab:rmse_x4_int_lidar}, Fig.~\ref{fig:results_int_x4} and Fig.~\ref{fig:results_lidar_int_x4}, we can find that the proposed approach outperforms other methods on all testing depth maps with smaller $RMSE$ results under non-linear (bi-cubic) degradation with noise and interval down-sampling degradation, which demonstrates the excellent generalization ability of the proposed framework on both synthesis and raw depth maps.

\vspace{-2mm}
\subsection{Weight sharing}
\vspace{-2mm}
As reported in Table.~\ref{tab:rmse_x4_int}, Table.~\ref{tab:rmse_x4_noise_bic} and Table.~\ref{tab:rmse_x4_int_lidar}, the proposed framework with weight-sharing among different sub-modules outperforms state-of-the-art methods, while it gets similar results with non-weight-sharing strategy. The last sub-module combines the outputs of previous sub-modules as input, hence, we use weight-sharing in other sub-modules except the last one. And the parameters of weight-sharing are only 40\% of parameters of non-weight-sharing ($K=5$), which makes the proposed framework lightweight and more flexible in comparison with other state-of-the-art methods.
\section{Conclusions}
In this paper, we have proposed an effective depth map super-resolution method that accounts for real-world degradation processes of different types of physical depth sensors. We have envisaged the employment of our new method to super-resolve depth maps captured by commodity depth sensors such as Microsoft Kinect and Lidar. We analyze two different LR depth map simulation schemes: non-linear downsampling and interval downsampling. Furthermore, we have devised a channel attention based iterative residual learning framework to address real world depth map super-resolution. Extensive experiments across different benchmarks have demonstrated the superiority of our proposed approach over the state-of-the-art.


{\small
\bibliographystyle{ieee_fullname}
\bibliography{cvpr2020_final}
}

\end{document}